\pdfoutput=1

\documentclass[final]{cvpr}

\usepackage{times}
\usepackage{epsfig}
\usepackage{graphicx}
\usepackage{amsmath}
\usepackage{amssymb}

\usepackage{color}
\usepackage{xcolor}
\usepackage{mathtools}
\usepackage{verbatim}
\usepackage[utf8x]{inputenc}
\usepackage{caption}
\usepackage{subfigure}
\usepackage{booktabs}
\usepackage{diagbox}
\definecolor{color_memory_branch}{RGB}{185, 218, 165}
\definecolor{color_query_branch}{RGB}{177, 206, 234}

\usepackage[pagebackref=true,breaklinks=true,colorlinks,bookmarks=false]{hyperref}

\def\cvprPaperID{7848} 
\def\confYear{CVPR 2021}

\begin{document}

\title{STMTrack: Template-free Visual Tracking with Space-time Memory Networks}

\author{Zhihong Fu, Qingjie Liu\thanks{Corresponding author.},\ \ Zehua Fu, Yunhong Wang\\
{\tt\small \{fuzhihong, qingjie.liu, yhwang\}@buaa.edu.cn, zehua\_fu@163.com}
}

\maketitle

\pagestyle{empty}  
\thispagestyle{empty} 

\begin{abstract}
Boosting performance of the offline trained siamese trackers is getting harder nowadays since the fixed information of the template cropped from the first frame has been almost thoroughly mined, but they are poorly capable of resisting target appearance changes.
Existing trackers with template updating mechanisms rely on time-consuming numerical optimization and complex hand-designed strategies to achieve competitive performance, hindering them from real-time tracking and practical applications.
In this paper, we propose a novel tracking framework built on top of a space-time memory network that is competent to make full use of historical information related to the target for better adapting to appearance variations during tracking.
Specifically, a novel memory mechanism is introduced, which stores the historical information of the target to guide the tracker to focus on the most informative regions in the current frame.
Furthermore, the pixel-level similarity computation of the memory network enables our tracker to generate much more accurate bounding boxes of the target.
Extensive experiments and comparisons with many competitive trackers on challenging large-scale benchmarks, OTB-2015, TrackingNet, GOT-10k, LaSOT, UAV123, and VOT2018, show that, without bells and whistles, our tracker outperforms all previous state-of-the-art real-time methods while running at 37 FPS.
The code is available at \url{https://github.com/fzh0917/STMTrack}.
\end{abstract}

\section{Introduction}
\begin{figure}[t]
    \centering
    \includegraphics[width=0.46\textwidth]{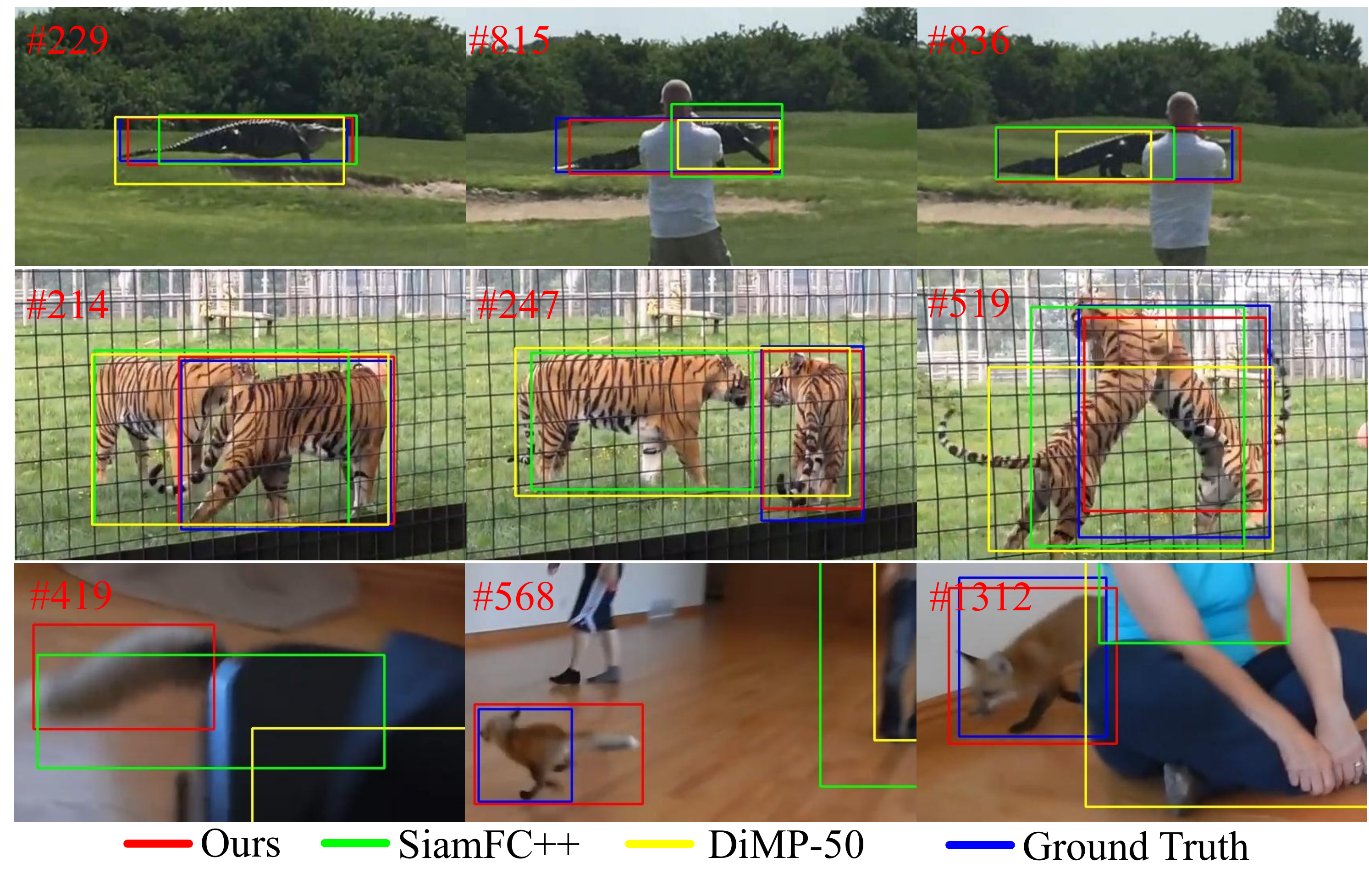}
    \vspace{-1.0em}
    \caption{Visualized comparisons of our method with representative trackers SiamFC++ \cite{xu2020siamfc++} and DiMP-50 \cite{bhat2019learning}. Our method can estimate more accurate target state when targets suffer from partial occlusions and non-rigid deformations.}
    \label{fig:demo_at_the_first_page_fig}
 \vspace{-1.0em}
\end{figure}

Visual object tracking is an essential task in computer vision with applications in various fields such as human-computer interactions~\cite{liu2012hand}, video surveillance~\cite{xing2010multiple}, and autonomous driving~\cite{lee2015road}.
Significant efforts have been devoted to address this problem, yet there is still a great gap to the practical applications due to the challenging factors such as occlusions, fast motions, and non-rigid deformations~\cite{fan2019lasot,otb2015,kristan2018sixth}, which urge us to develop trackers with strong adaptiveness and robustness.
\par
The goal of visual tracking is to locate an object in the subsequent frames of a video given its initial annotation in the first frame.
In recent years, with the advancements of deep learning techniques, deep trackers have dominated the tracking field, among which two methodologies are widely studied, and one popular methodology addresses object tracking as a similarity matching problem between the target template and the search frames in an embedding space offline trained.
The representative template-matching methods are siamese trackers~\cite{bertinetto2016fully,zhu2018distractor,zhang2019deeper,fan2019siamese,wang2019fast,chen2020siamese,li2018high,li2019siamrpn++,xu2020siamfc++,guo2020siamcar}.
These methods usually do not update the template and thus are hard to adapt to appearance changes caused by occlusions, non-rigid deformations, etc.
\par
To solve this problem, some trackers~\cite{bhat2019learning,danelljan2020probabilistic} are equipped with sophisticated template updating mechanisms and thus show stronger robustness than siamese trackers.
However, online template updating requires much more computational resources, which impends trackers from real-time tracking.
Furthermore, these customized updating strategies~\cite{guo2017learning,yang2018learning,zhu2018distractor,li2019gradnet,yu2020deformable,choi2019deep} introduce hyper-parameters that require tricky tuning.
\par
Note that when tracking moving objects humans remember their identities in visual working memory to maintain temporal continuity in a constantly changing environment~\cite{makovski2009role}.
This inspires us to develop a memory-based tracking model to take advantage of rich historical information of the object.
In contrast to previous works that strive to design template updating mechanisms to capture appearance variations of the object, our model predicts the state of the object from its historical information stored in the memory network, thus avoiding of using the template and updating it.
Thus we call our model a template-free method.
Furthermore, our tracker computes pixel-level similarities to locate the target, making it more robust to partial occlusions and non-rigid deformations than those using feature-map-level cross correlation.
\figref{fig:demo_at_the_first_page_fig} depicts this advantage of our tracker (Refer to the section 1.1 of the supplemental material for quantitative comparisons).
\par
The proposed method is evaluated on six benchmarks: OTB-2015, TrackingNet, LaSOT, GOT-10k, UAV123, and VOT2018 and surpasses all state-of-the-art real-time approaches while running at 37 FPS.
Notably, it achieves 80.3 success (AUC) on the challenging TrackingNet dataset, outperforming the previous best real-time method by $4.5\%$.
It also sets new state-of-the-art performance on the performance-saturated OTB-2015 dataset.

Summarily, the main contributions of this work are fourfold.
\vspace{-0.5em}
\begin{itemize}
\item{We propose a novel end-to-end memory-based tracking framework, which not only is as simple and efficient as the offline trained siamese networks, but also has strong adaptiveness ability as the sophisticated template updating strategies.}
\vspace{-0.5em}
\item{The proposed tracking framework deviates from the original evolving path of template-based tracking, and it could inspire more space-time-memory-based trackers to be developed in the future.}
\vspace{-0.5em}
\item{A novel memory mechanism based on pixel-level similarity computation is introduced for visual tracking, which enables our tracker to have stronger robustness and to generate much more accurate target bounding boxes than many previous high-performance methods that use feature-map-level cross correlation.}
\vspace{-0.5em}
\item Our proposed tracker outperforms all state-of-the-art real-time approaches on OTB-2015, TrackingNet, LaSOT, and GOT-10k, while running in real-time at 37 FPS, which demonstrates the superiority of the framework.
\end{itemize}

\begin{figure*}[t]
    \centering
    \includegraphics[width=0.98\textwidth]{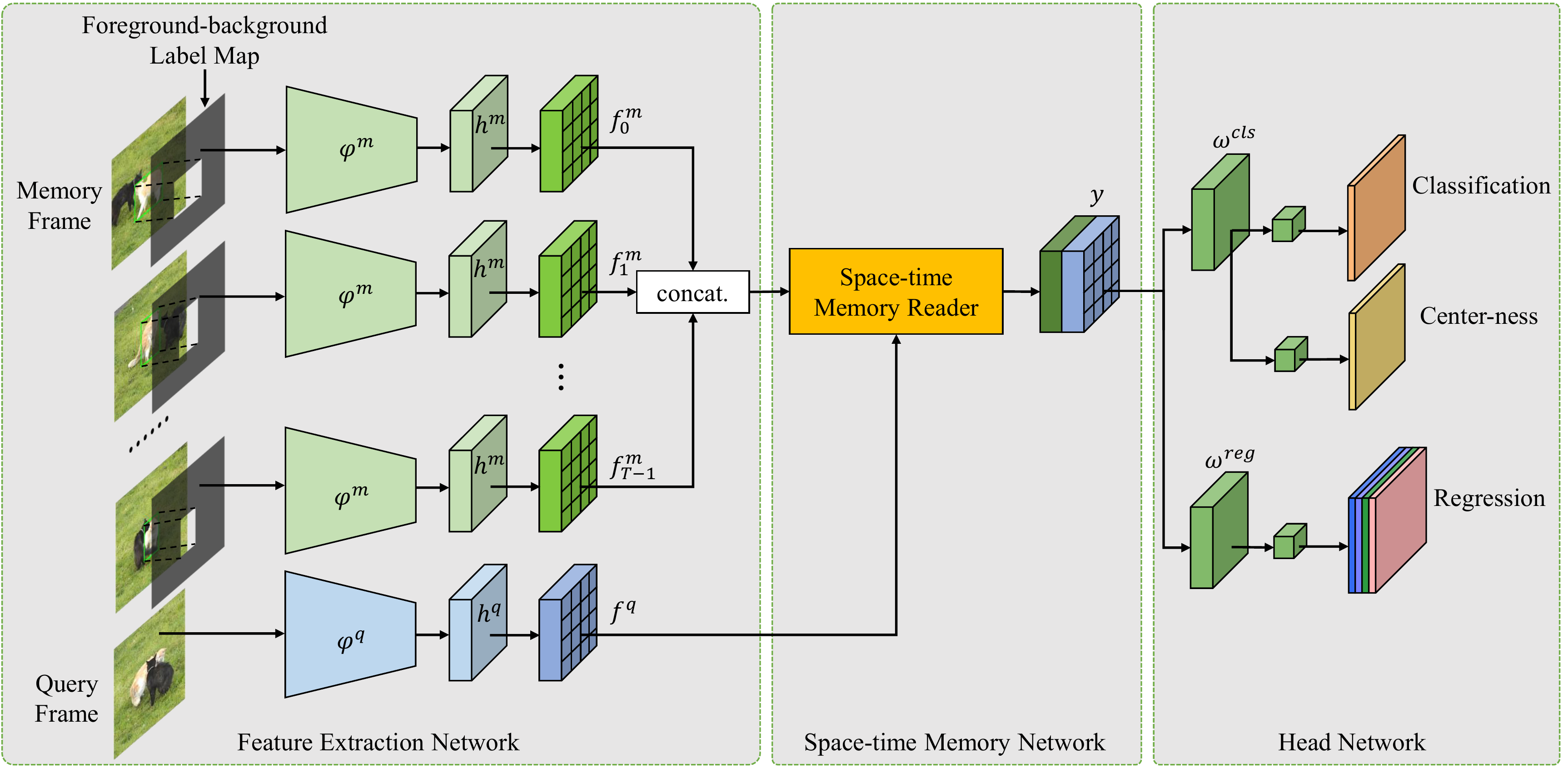}
    \vspace{-1.0em}
    \caption{The architecture of our proposed method. The left part is the feature extraction network that consists of a memory branch (displayed in \textcolor{color_memory_branch}{light green}) and a query branch (displayed in \textcolor{color_query_branch}{light blue}). The memory branch takes both memory frames and corresponding foreground-background label maps as inputs. \lq\lq{concat.}\rq\rq{} denotes the concatenation operation along the temporal dimension. The middle part is the space-time memory network that retrieves the target information from multiple memory frames for the target localization in the query frame. The right side is the head network for the foreground-background classification and the target bounding box regression of the query frame.}
    \label{fig:architecture}
 \vspace{-1.0em}
\end{figure*}

\section{Related Work}
\subsection{Siamese Trackers}
Siamese networks~\cite{bromley1993signature,zagoruyko2015learning,chopra2005learning} have attracted significant attention in recent years and been very popular in the tracking community~\cite{bertinetto2016fully,zhu2018distractor,zhang2019deeper,fan2019siamese,wang2019fast,chen2020siamese,li2018high,li2019siamrpn++,xu2020siamfc++,guo2020siamcar} .
Siamese trackers treat the visual object tracking task as a matching problem.
During inference, a template is cropped from the first frame and matched to the search regions in the current frame to achieve tracking.
They have excellent performance and real-time tracking speed when running in many routine tracking scenarios.
However, their vulnerability becomes evident when the targets suffer from drastic appearance changes, non-rigid deformations, and partial occlusions.
Unlike these approaches, our proposed work makes full use of historical multiple-frame information in the tracking process, which can greatly improve the robustness of the model in those challenging scenarios.
\subsection{Template Updates}
To cope with challenging factors in tracking processing, updating the template is critical to adapt the tracker to target variations.
\par
DSiam~\cite{guo2017learning} proposes a dynamic siamese network with a fast transformation learning model to enable effective template updating and cluttered background suppression.
In~\cite{zhu2018distractor}, a distractor-aware module is designed to transfer the general embedding in siamese networks to the specific target domain of the current video.
Observing that the absolute values of gradients at locations where objects are occluded or distracted from similar objects are prone to be higher, GradNet~\cite{li2019gradnet} integrates backward gradients into the initial template for augmenting the discriminative ability of the template.
These methods explicitly select informative frames and use customized strategies to update the template.
Different from these explicit template updating strategies, we propose to store the historical information of the target in memory networks and retrieve them as needed.

\subsection{Memory Networks}
Memory networks were first proposed to solve document Q\&A~\cite{weston2014memory,sukhbaatar2015end,miller2016key,kumar2016ask} in the Natural Language Processing field.
They are common neural networks equipped with external memory components to read and write historical information.
Recently, memory networks have shown significant performance improvement in few shot learning~\cite{santoro2016meta,lu2020video}, video object segmentation~\cite{oh2019video,lu2020video}, \etc
\par
Memory networks have been also introduced into visual tracking, a typical one is MemTrack~\cite{yang2018learning}.
It uses a memory network to read a residual template during tracking and then combines it with the initial template to yield a synthetic template that is used as an updated representation of the target.
Although MemTrack~\cite{yang2018learning} uses a large amount of historical information in the tracking process, the memory reading operation controlled by a LSTM may lose useful information.
The overall performance of the tracker is greatly affected by the learning quality of the LSTM controller.
\par
In our work, the retrieval of historical information is determined by the current frame itself, therefore it can obtain all it needs, adaptively.

\section{Proposed Method}
In this section, we will describe the proposed tracking framework in detail.
First, we will introduce an overview of the framework in \secref{subsec:architecture}.
Then, we will give an account of each module of the whole framework one by one from \secref{subsec:feature-extraction-network} to \secref{subsec:head-network}.
Last, we will present the online tracking process of the framework in \secref{subsec:inference-phase}.

\subsection{Architecture}\label{subsec:architecture}
As shown in \figref{fig:architecture}, the framework can be divided into three parts, a feature extraction network, a space-time memory network, and a head network.
The feature extraction network consists of a memory branch (in \textcolor{color_memory_branch}{light green}) and a query branch (in \textcolor{color_query_branch}{light blue}).
The memory branch takes both memory frames and corresponding foreground-background label maps (will be explained in the next section) as inputs, while the input of the query branch is only a single query frame.
In this work, the memory frames are multiple historical frames, and the query frame is the current frame in a tracking sequence.
After the feature extraction, the space-time memory network retrieves information related to the target from features of all memory frames, generating a synthetic feature map to classify the target from backgrounds and to predict the target bounding box for the query frame.

\subsection{Feature Extraction Network}\label{subsec:feature-extraction-network}
Here we describe the feature extraction procedures of the memory branch and the query branch, respectively.
\par
\textbf{Memory Feature Extraction.}
The inputs of the memory branch are $T$ memory frames $m$ (each frame is $m_i$) and $T$ foreground-background label maps $c$ (each label map is $c_i$), where $c$ is to ensure that the memory backbone $\varphi^{m}$ learns the consistency of the real target characteristics rather than distractors and cluttered background information.
Specifically, we label each pixel with $1$ within the corresponding ground truth bounding box and $0$ elsewhere for each memory frame $m_i$.
Then, we adopt the first convolutional layer of $\varphi^{m}$ (represented by $\varphi^{m}_{0}$) and an extra convolutional layer $g$ to map $m$ and $c$ to a same embedding space, respectively.
After that, we add $\varphi^{m}_{0}(m)$ and $g(c)$ element-wise, then input the sum to the latter layers of $\varphi^{m}$ to generate $T$ memory feature maps (denoted as $f^{m}$, each memory feature map is $f^{m}_{i}$).
The feature dimensionality of $f^{m}$ is then reduced to 512 by a non-linear convolutional layer (denoted as $h^{m}$):
\begin{equation}\label{eq:backbone_m}
\begin{split}
f^{m}_{i}=h^{m}(\varphi^{m}_{\gamma}(\varphi^{m}_{0}(m_{i}) \oplus g(c_{i})))
\end{split}
\end{equation}
where $f^{m}_{i} \in \mathbb{R}^{C \times H \times W}$, $\varphi^{m}_{\gamma}$ represents all layers of $\varphi^{m}$ except the first layer, and \lq\lq{$\oplus$}\rq\rq{} is element-wise addition.
\par
\textbf{Query Feature Extraction.}
Different from the memory branch, the query branch takes a query frame $q$ as input and produces a feature map $\varphi^{q}(q)$.
Similar to the memory branch, the feature dimensionality of $\varphi^{q}(q)$ is also reduced to 512 by a non-linear convolutional layer (denoted as $h^{q}$):
\begin{equation}\label{eq:backbone_q}
\begin{split}
f^{q}=h^{q}(\varphi^{q}(q))
\end{split}
\end{equation}
where $f^{q} \in \mathbb{R}^{C \times H \times W}$.
\par
Note that the two backbones $\varphi^{m}$ and $\varphi^{q}$ share the same network architecture but have different parameters.
An ablation study on whether sharing one backbone can be seen in \secref{subsec:ablation-study}.
\par

\subsection{Space-time Memory Network}\label{subsec:space-time-memory-reader}
\begin{figure}[t]
    \centering
    \includegraphics[width=0.46\textwidth]{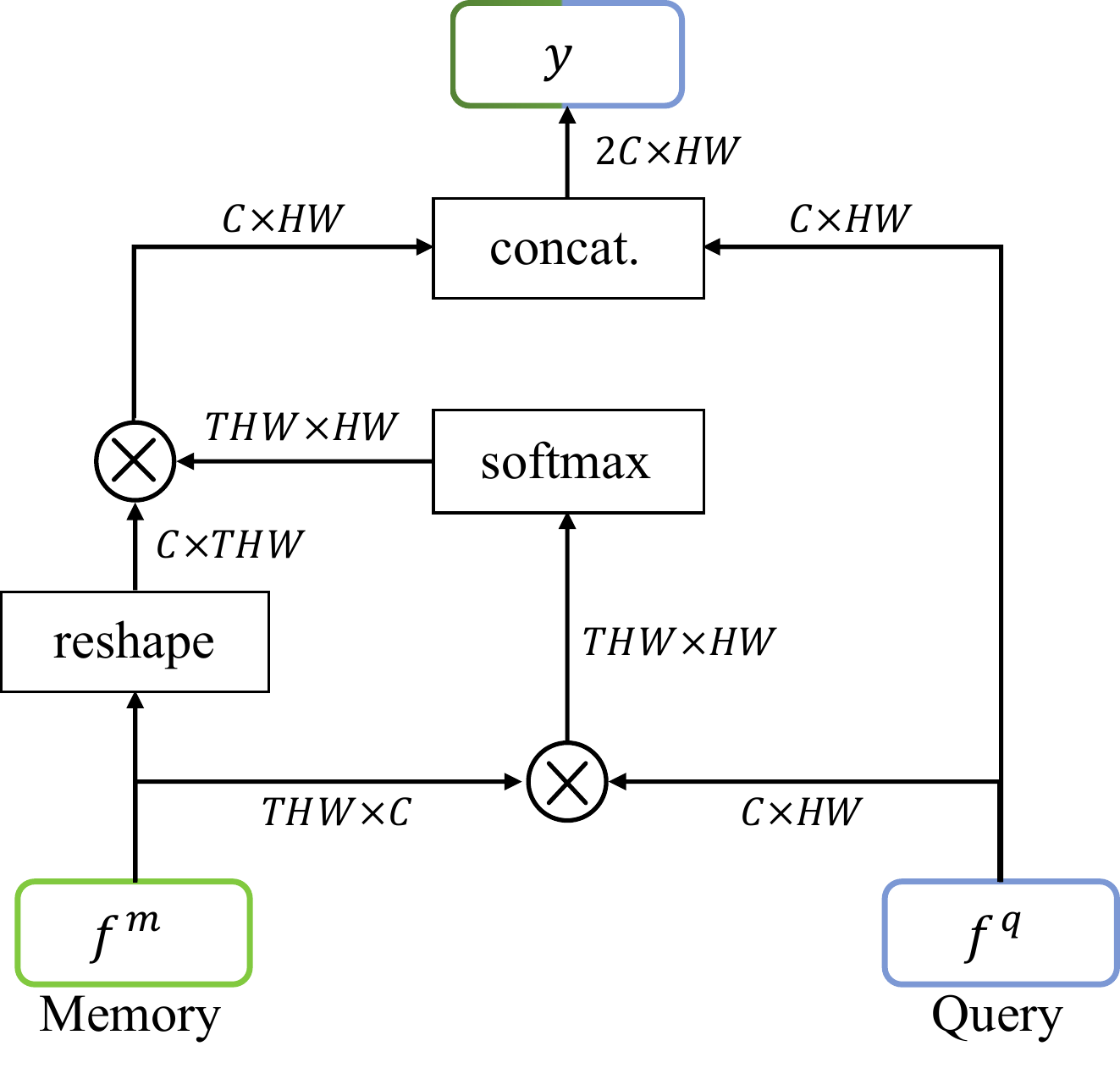}
    \vspace{-1.0em}
    \caption{The space-time memory reader. Here $f^{m} \in \mathbb{R}^{T \times C \times H \times W}$ and $f^{q} \in \mathbb{R}^{C \times H \times W}$, where $T$ is the number of memory frames, $C$, $H$ and $W$ represent the feature dimensionality, the height, and the width of the feature map, respectively. For the convenience of matrix multiplication in math, we reshape $f^{m}$ from $T \times C \times H \times W$ to $THW \times C$, and reshape $f^{q}$ from $C \times H \times W$ to $C \times HW$, thus here $THW = T \times H \times W$ and $HW = H \times W$. the operator \lq\lq{$\otimes$}\rq\rq{} denotes matrix multiplication, and \lq\lq{concat.}\rq\rq{} denotes the concatenation operation along the channel dimension.}
    \label{fig:memory_reader_fig}
 \vspace{-1.0em}
\end{figure}

As illustrated in \figref{fig:memory_reader_fig}, we first compute the similarities between every pixel of $f^{m}$ and every pixel of $f^{q}$ to obtain a similarity matrix $w \in \mathbb{R}^{THW \times HW}$.
Inspired by \cite{wang2018non}, we expect the similarity computation to apply the gaussian function.
Thus, we normalize $w$ with a \verb!softmax! function.
Taking one element $w_{ij}$ for example, we can formally denote $w_{ij}$ as:
\begin{equation}\label{eq:memory-read-eq1}
\begin{split}
w_{ij}=\frac{\exp\left[\left(f^{m}_{i \cdot} \odot f^{q}_{\cdot j}\right) / s\right]}{\sum\limits_{\forall k}\exp\left[\left(f^{m}_{k \cdot} \odot f^{q}_{\cdot j}\right) / s\right]}
\end{split}
\end{equation}
where $i$ is the index of each pixel on $f^m \in \mathbb{R}^{THW \times C}$, $j$ is the index of each pixel on $f^q \in \mathbb{R}^{C \times HW}$, and the binary operator $\odot$ denotes vector dot-product.
Here $s$ is a scaling factor to prevent the \verb!exp! function from overflowing numerically.
Following \cite{Vaswani2017AttentionIA}, we set $s$ to $\sqrt{C}$, where $C$ is the feature dimensionality of $f^{m}$.

Then, treating $w$ as a soft weight map, we multiply $f^{m}$ by $w$.
Because $f^{m}$ stores all historical memory information related to the target, according to the needs of the query frame itself, the target information stored in $f^{m}$ is adaptively retrieved.
Obviously, the readout information is a feature map as the same size as $f^{q}$.
Therefore, we concatenate the readout information and the query feature map $f^{q}$ along the channel dimension to generate the final synthetic feature map $y$.
Formally, for the $i$-th element of $y$, the space-time memory read operation can be denoted as:
\begin{equation}\label{eq:memory-read}
\begin{split}
y_{i}=\text{concat}\left(f^{q}_{i}, (f^{m})^{T}_{i} \otimes w \right)
\end{split}
\end{equation}
where $(f^{m})^{T} \in \mathbb{R}^{C \times THW}$ is the transpose of $f^{m}$, and the $\verb!concat!(\cdot, \cdot)$ function represents the concatenation operation.

At a first glance, the working mechanism of the memory read operation is similar to the non-local self-attention \cite{wang2018non}.
A representative example of deploying the non-local self-attention \cite{wang2018non} in visual tracking is AlphaRefine \cite{yan2020alpha}, the winner of the real-time tracking challenge VOT-RT2020 \cite{Kristan2020a}, that uses a non-local block to augment the response map generated by a pixel-wise correlation since the longer-range dependencies can produce more precise target boundary decision information.
Differently, the purpose of designing the space-time memory reader in our proposed framework, however, is to retrieve the target information from multiple memory frames by taking the similarity matrix as soft weights, instead of computing the non-local self-attention for each pixel pair in a feature map.
\par
Particularly, different from STMVOS~\cite{miller2016key} and GraghMemVOS~\cite{oh2019video} in video object segmentation, our method does not divide the features extracted by $\varphi^{m}$ and $\varphi^{q}$ into keys and values, but directly uses $f^{m}$ and $f^{q}$ to locate the target. The motivation is that, $f^{m}$ itself happens to provide adequate target information to find the exposed parts of the target when it suffers from partial occlusions in the query frame.
This difference makes the space-time memory network more suitable for the single object tracking task.

\subsection{Head Network}\label{subsec:head-network}
Inspired by the phenomenon that the one-stage anchor-free detector \cite{tian2019fcos} has achieved better performance and has fewer parameters than the one-stage anchor-based method \cite{lin2017focal} in object detection, we design an anchor-free head network that contains a classification branch to classify the target from backgrounds and an anchor-free regression branch to directly estimate the target bounding box.
\par
To be specific, first, we encode $y$ with a lightweight classification convolutional network $\omega^{cls}$ to integrate $f^{q}$ and the retrieved information from $f^{m}$ to adapt to the classification task.
Then, a linear convolutional layer with $1 \times 1$ kernel is used to reduce the dimensionality of the output of $\omega^{cls}$ to 1, producing the final classification response map $R^{cls} \in \mathbb{R}^{1 \times H \times W}$.
\par
Moreover, we observe that the positive samples near the target boundary tend to predict low-quality target bounding boxes.
Therefore, a sub-branch is forked after $\omega^{cls}$ to generate a center-ness response map $R^{ctr} \in \mathbb{R}^{1 \times H \times W}$, as illustrated in the right part of \figref{fig:architecture}.
During inference, $R^{cls}$ is multiplied by $R^{ctr}$ to suppress the classification confidence scores of pixels away from the target center.
\par
In the regression branch, we pass $y$ to another lightweight regression convolutional network $\omega^{reg}$ and then reduce the dimensionality of the outputted features to 4 to generate a regression response map $R^{reg} \in \mathbb{R}^{4 \times H \times W}$ for the target bounding box estimation.
\par
We recommend readers to refer to \cite{xu2020siamfc++} for more details about the encodings and the training objectives of $R^{cls}$, $R^{ctr}$, and $R^{reg}$.

\subsection{Inference Phase}\label{subsec:inference-phase}
Our space-time memory network is flexible so that the number of used memory frames (\ie the memory size) during inference is independent of the number of memory frames during training (See \secref{subsec:ablation-study} for the impact of different number of memory frames on performance in the two phases).
In this work, for the current frame $F_{t}$, we select $N$ memory frames from all historical frames (\ie frame $F_{1}$ to frame $F_{t-1}$) as memory frames for rich appearance information and strong generalization ability.
From the perspective of existing works \cite{voigtlaender2020siam,oh2019video}, experiences, and intuitions, target information from the first frame and the previous frame plays an important role for the target localization in the current frame.
Specifically, the target from the first frame provides the most reliable information, while the tracked target from the previous frame has the most similar appearance to the target in the current frame.
Therefore, for the current frame $F_{t}$, the memory frames hold the first frame $F_1$, the previous frame $F_{t-1}$ and other $N-2$ frames $F_{\tau_1},F_{\tau_2},\cdots,F_{\tau_{\left({N-2}\right)}}$ sampled following the methodology: splitting all historical frames into $N-2$ segments, and choosing one representative frame from each segment for the best balance between the target domain adaptation, underfitting, and the time cost.
Formally, the sampling method can de described as:
\begin{equation}\label{eq:inference-sampling}
\begin{split}
\tau_{i}=\left\lfloor \left\lfloor \frac{t-1}{N-2} \right\rfloor \times \left(i + \Delta_{i}\right) \right\rfloor
\end{split}
\end{equation}
Here, $i \in \{1,2,\cdots,N-2\}$, and $\Delta_{i} \in [0, 1)$ is the offset of the representative frame in the $i$-th segment.
For the first $N$ frames, we set all historical frames (\ie $F_{1},F_{2},\cdots,F_{N-1}$) as memory frames.
In our experiments, $N$ is set to $6$, and we simply set $\left\{\Delta_{i}=\frac{1}{2} | 1 \le i \le N-2 \right\}$.

For each frame in the whole tracking process, after obtaining $R^{cls}$, $R^{ctr}$, and $R^{reg}$, the postprocessing is the same as \cite{xu2020siamfc++}.

\section{Experiments}
Our tracker is implemented in Python using \textit{PyTorch} framework, which runs at 37\footnote{The running speed is calculated by the GOT-10k evaluation server.} FPS on an NVIDIA Tesla V100 GPU.
We evaluate our tracker on benchmarks: OTB-2015 \cite{otb2015}, TrackingNet \cite{muller2018trackingnet}, GOT-10k \cite{huang2019got}, LaSOT \cite{fan2019lasot}, UAV123 \cite{mueller2016benchmark}, and VOT2018~\cite{kristan2018sixth}.

\subsection{Training Dataset}\label{subsec:training-dataset}
We adopt TrackingNet \cite{muller2018trackingnet}, the \textit{training} set of LaSOT \cite{fan2019lasot} and GOT-10k \cite{huang2019got}, ILSVRC VID~\cite{russakovsky2015imagenet}, ILSVRC DET~\cite{russakovsky2015imagenet}, and COCO \cite{lin2014microsoft} as our training dataset except the GOT-10k benchmark.
We sample $T$ ($T=3$ in this paper) frames within the maximum frame index gap 100 and adopt random affine transformations to increase data diversity for training video sequences.
To be specific, the translation is randomly performed from $-0.2S$ to $0.2S$ and the resizing scale varies between $\frac{1}{1+r}$ and $1+r$ with $r=0.3$.
Here $S$ is the cropping size, and we set $S$ to the scale of 4 times the target bounding box for mining as much contextual information as possible and enhancing the discriminative ability of the model.
We apply the above data augmentation strategy with different parameter settings to generate a synthetic video sample for ILSVRC DET \cite{russakovsky2015imagenet} and COCO \cite{lin2014microsoft}.
Note that, different from siamese trackers, our method does not have to keep the same target scale for a training sample.
For each frame in a training sample, taking the target center as the center, we crop a square image patch with side length $S$ from the original frame and resize the cropped image patch to $289 \times 289$ to be the input of the model.

\subsection{Implementation Details}
\textbf{Model Settings.}
We adopt GoogLeNet~\cite{szegedy2015going} as our backbone $\varphi^{m}$ and $\varphi^{q}$.
The classification convolutional network $\omega^{cls}$ and the regression convolutional network $\omega^{reg}$ are both comprised of seven convolutional layers.
Each convolutional layer in $\omega^{cls}$ and $\omega^{reg}$ is followed by a \verb!ReLU! activation function.
\par
\textbf{Optimization and Training Strategies.}
Our tracker is trained with SGD optimizer for 20 epochs.
There are 300,000 samples per epoch, and the mini-batch size is set to 64.
The whole training phase takes about 27 hours to converge with four NVIDIA Tesla V100 GPUs.
For the GOT-10k benchmark, we set the number of samples per epoch to 150,000 and the mini-batch size to 32.
The momentum and the weight decay rate are set to 0.9 and $1 \times 10^{-4}$, respectively.
We freeze all convolutional layers of $\varphi^{m}$ and $\varphi^{q}$ in the first 10 epochs and unfreeze all convolutional layers of stage 3 and 4~\cite{szegedy2015going} of $\varphi^{m}$ and $\varphi^{q}$ for the other epochs.
The learning rate increases from $1 \times 10^{-2}$ to $8 \times 10^{-2}$ with a warmup technology at the first epoch and decreases from $8 \times 10^{-2}$ to $1 \times 10^{-6}$ with a cosine annealing learning rate schedule for the other epochs.

\subsection{Ablation Study}\label{subsec:ablation-study}
In this section, we conduct a comprehensive ablation study for the proposed tracker.
\par
\textbf{Should the Tracker Share one Backbone?}
First, we analyze whether the tracker should share one backbone between the memory branch and the query branch like most siamese networks. As shown in \tabref{tab:ablation_study_backbone_sharing}, from the top two rows, we observe that sharing one backbone can improve the average overlap (AO) by $2.3\%$ if we do not use foreground-background label maps.
However, the comparison of the bottom two rows in \tabref{tab:ablation_study_backbone_sharing} and comparisons on other benchmarks in \tabref{tab:ablation_study_backbone_sharing2} show that the performance of using different backbones is more superior than sharing one.

\textbf{Should the Tracker Use Foreground-background Label Maps in the Input of the Memory Branch?}
Second, we discuss the necessity to use foreground-background label maps in the input of the memory branch.
By comparing the first row with the third row and comparing the second row with the last row in \tabref{tab:ablation_study_backbone_sharing}, we notice that the performance will be improved by $2.9\%$ and $7.4\%$ with the same backbone and different backbones, respectively.
It indicates that foreground-background label maps are crucial to the memory mechanism in this framework.

\textbf{What is the Best Number of Reference Frames in the Memory?}
Last but not least, the number of reference frames in the memory (\ie the memory size) is a key factor for memory networks.
During training, the number of reference frames affects the learning qualities of two backbones.
The more reference frames, the more target patterns can be trained, but the more frames are similar to the current frame.
In that case, the network tends to compare the most similar image pairs, rather than learning to compute the similarities between the current frame and frames with clutter backgrounds or partially occluded targets.
We verify the impact in training with different memory size settings on the GOT-10k benchmark.
\tabref{tab:ablation_study_num_frames_for_training} shows that using 3 reference frames in a training sample brings the best performance in terms of average overlap (AO) metric.
\par
During inference, the memory size not only affects the performance, but also significantly determines the running speed.
We conduct a group of experiments on TrackingNet to analyze the impact of the memory size on the performance and the speed.
As listed in \tabref{tab:ablation_study_num_frames_for_testing}, the most suitable memory size is 6 for this work.
Besides, the more reference frames does not produce the better performance.
We speculate that there are two main reasons: the one is overfitting, and the another one is too many low-quality memory frames affect tracking results that further products more inferior memory frames.

\begin{table}[!tbp]
 \centering
 \caption{\label{tab:ablation_study_backbone_sharing}Ablation study on the GOT-10k benchmark.  Here the variable \lq\lq{share}\rq\rq{} denotes whether the network should share the backbone between the memory and the query branch, and the variable \lq\lq{fb\_label}\rq\rq{} represents whether the network should use foreground-background label maps in the memory branch input.}
 \vspace{-1.0em}
 \begin{tabular}{cc|ccc}
  \toprule
  share & fb\_label & \textbf{AO} & $\textbf{\text{SR}}_{0.5}$ & $\textbf{\text{SR}}_{0.75}$ \\
  \midrule
  \checkmark & - & 0.591 & 0.662 & 0.507 \\
  - & - & 0.568 & 0.638 & 0.480 \\
  \checkmark & \checkmark & 0.620 & 0.713 & 0.538 \\
  - & \checkmark & \textcolor{red}{\textbf{0.642}} & \textcolor{red}{\textbf{0.737}} & \textcolor{red}{\textbf{0.579}} \\
  \bottomrule
 \end{tabular}
\end{table}

\begin{table}[!tbp]
 \centering
 \caption{\label{tab:ablation_study_backbone_sharing2}The same ablation study as \tabref{tab:ablation_study_backbone_sharing} on OTB-2015, TrackingNet, and LaSOT. The tracker is evaluated by success (AUC) metric.}
 \vspace{-1.0em}
 \begin{tabular}{cc|cccc}
  \toprule
  share & fb\_label & \textbf{OTB-2015} & \textbf{TrackingNet} & \textbf{LaSOT} \\
  \midrule
  \checkmark & \checkmark & 0.702 & 79.7 & 0.593 \\
  - & \checkmark & \textcolor{red}{\textbf{0.719}} & \textcolor{red}{\textbf{80.3}} & \textcolor{red}{\textbf{0.606}} \\
  \bottomrule
 \end{tabular}
\end{table}

\begin{table}[!tbp]
 \centering
 \caption{\label{tab:ablation_study_num_frames_for_training}The performance on GOT-10k with different number of reference frames in training. Here \textbf{AO} is the average overlap metric.}
 \vspace{-1.0em}
 \begin{tabular}{c|cccc}
  \toprule
  \textbf{\#} & \textbf{1} & \textbf{2} & \textbf{3} & \textbf{4} \\
  \midrule
  \textbf{AO} & 0.629 & 0.624 & \textcolor{red}{\textbf{0.642}} & 0.627 \\
  \bottomrule
 \end{tabular}
\end{table}

\begin{table}[!tbp]
 \centering
 \caption{\label{tab:ablation_study_num_frames_for_testing}The performance in terms of success (AUC) metric on TrackingNet with different number of reference frames in the inference phase.}
 \vspace{-1.0em}
 \begin{tabular}{c|cccccc}
  \toprule
  \textbf{\#} & \textbf{1} & \textbf{2} & \textbf{4} & \textbf{6} & \textbf{8} & \textbf{ALL}  \\
  \midrule
  \textbf{Success} & 79.1 & 79.3 & 80.2 & \textcolor{red}{\textbf{80.3}} & 80.2 & 79.8 \\
  \textbf{FPS} & 43.0 & 26.6 & 29.3 & 28.6 & 22.7 & 6.5 \\
  \bottomrule
 \end{tabular}
 \vspace{-1.0em}
\end{table}

\subsection{Comparison with the state-of-the-art}
\textbf{OTB-2015.}
OTB-2015 \cite{otb2015} is a classical benchmark in visual object tracking, containing 100 short-term videos with 590 frames per video on average.
We report the results on OTB-2015.
It is known to have tended to saturation over recent years.
Still, as shown in \tabref{tab:otb2015_tab}, our approach surpasses the previous best performance trackers by $0.4\%$ in terms of success (AUC) metric, setting a new state-of-the-art performance on this dataset.

\begin{table}[!tbp]
 \centering
 \caption{\label{tab:otb2015_tab}A success (AUC) performance list on OTB-2015 for a comprehensive comparison of our tracker with competitive trackers published in recent years. The best three results are highlighted in \textcolor{red}{\textbf{red}}, \textcolor{blue}{\textbf{blue}}, and \textcolor{green}{\textbf{green}}, respectively. Trackers are ranked from top to bottom and left to right according the \textbf{Success} values.}
 \vspace{-1.0em}
 \begin{tabular}{cc|cc}
  \toprule
  \textbf{Tracker} & \textbf{Success} & \textbf{Tracker} & \textbf{Success} \\
  \midrule
  \textbf{Ours} & \textcolor{red}{\textbf{0.719}} & SiamRPN++ \cite{li2019siamrpn++} & 0.696 \\
  DROL \cite{zhou2020discriminative} & \textcolor{blue}{\textbf{0.715}} & KYS \cite{bhat2020know} & 0.695 \\
  RPT \cite{ma2020rpt} & \textcolor{blue}{\textbf{0.715}} & MCCT \cite{wang2018multi} & 0.695 \\
  CGACD \cite{du2020correlation} & \textcolor{green}{\textbf{0.713}} & GFS-DCF \cite{xu2019joint} & 0.693 \\
  SiamAttn \cite{yu2020deformable} & 0.712 & ASRCF \cite{dai2019visual} & 0.692 \\
  DCFST \cite{zhenglearning} & 0.709 & PGNet \cite{liaopg} & 0.691 \\
  UPDT \cite{bhat2018unveiling} & 0.702 & RPCF \cite{sun2019roi} & 0.690 \\
  DRT \cite{sun2018correlation} & 0.699 & SPM \cite{wang2019spm} & 0.687 \\
  SiamCAR \cite{guo2020siamcar} & 0.697 & DiMP-50 \cite{bhat2019learning} & 0.684 \\
  PrDiMP-50 \cite{danelljan2020probabilistic} & 0.696 & Ocean \cite{zhang2020ocean} & 0.684 \\
  SiamBAN \cite{chen2020siamese} & 0.696 & SiamFC++ \cite{xu2020siamfc++} & 0.683 \\
  \bottomrule
 \end{tabular}
\end{table}

\begin{table}[!tbp]
 \centering
 \caption{\label{tab:trackingnet_tab}A performance comparison of our tracker with other competitive approaches on the test split of TrackingNet. Trackers are ranked from top to bottom according the \lq\lq{\textbf{Suc.}}\rq\rq{} values. \lq\lq{\textbf{Suc.}}\rq\rq{}, \lq\lq{\textbf{Prec.}}\rq\rq{}, and \lq\lq{\textbf{Norm. Prec.}}\rq\rq{} are abbreviations for success (AUC), precision, and normalized precision, respectively. The best three results are highlighted in \textcolor{red}{\textbf{red}}, \textcolor{blue}{\textbf{blue}}, and \textcolor{green}{\textbf{green}}, respectively.}
 \vspace{-1.0em}
 \begin{tabular}{cccc}
  \toprule
  \textbf{Tracker} & \textbf{Suc.} & \textbf{Prec.} & \textbf{Norm. Prec.}\\
  \midrule
  \textbf{Ours} & \textcolor{red}{\textbf{80.3}} & \textcolor{red}{\textbf{76.7}} & \textcolor{red}{\textbf{85.1}} \\
  PrDiMP-50 \cite{danelljan2020probabilistic} & \textcolor{blue}{\textbf{75.8}} & 70.4 & 81.6 \\
  FCOS-MAML \cite{wang2020tracking} & \textcolor{green}{\textbf{75.7}} & - & \textcolor{blue}{\textbf{82.2}} \\
  SiamFC++ \cite{xu2020siamfc++} & 75.4 & \textcolor{green}{\textbf{70.5}} & 80.0 \\
  SiamAttn \cite{yu2020deformable} & 75.2 & - &  \textcolor{green}{\textbf{81.7}} \\
  DCFST-50 \cite{zhenglearning} & 75.2 & 70.0 & 80.9 \\
  DROL \cite{zhou2020discriminative} & 74.6 & \textcolor{blue}{\textbf{70.8}} &  \textcolor{green}{\textbf{81.7}} \\
  KYS \cite{bhat2020know} & 74.0 & 68.8 & 80.0 \\
  DiMP-50 \cite{bhat2019learning} & 74.0 & 68.7 & 80.1 \\
  SiamRPN++ \cite{li2019siamrpn++} & 73.3 & 69.4 & 80.0 \\
  D3S \cite{lukezic2020d3s} & 72.8 & 66.4 & 76.8 \\
  CGACD \cite{du2020correlation} & 71.1 & 69.3 & 80.0 \\
  GlobalTrack \cite{huang2020globaltrack} & 70.4 & 65.6 & 75.4 \\
  ATOM \cite{danelljan2019atom} & 70.3 & 64.8 & 77.1 \\
  \bottomrule
 \end{tabular}
 \vspace{-1.0em}
\end{table}

\begin{table}[!tbp]
 \centering
 \caption{\label{tab:got10k_tab}A performance comparison of our tracker with other competitive approaches on the test split of GOT-10k in terms of average overlap (AO) and success rates (SR) at threshold 0.5 and 0.75. The best three results are highlighted in \textcolor{red}{\textbf{red}}, \textcolor{blue}{\textbf{blue}}, and \textcolor{green}{\textbf{green}}, respectively. Trackers are ranked from top to bottom according the \textbf{AO} values.}
 \vspace{-1.0em}
 \begin{tabular}{cccc|c}
  \toprule
  \textbf{Tracker} & \textbf{AO} & $\textbf{\text{SR}}_{0.5}$ & $\textbf{\text{SR}}_{0.75}$ & \textbf{FPS}\\
  \midrule
  \textbf{Ours} & \textcolor{red}{\textbf{0.642}} & \textcolor{green}{\textbf{0.737}} & \textcolor{red}{\textbf{0.575}} & 37 \\
  KYS \cite{bhat2020know} & \textcolor{blue}{\textbf{0.636}} & \textcolor{red}{\textbf{0.751}} & \textcolor{green}{\textbf{0.515}} & 20 \\
  PrDiMP-50 \cite{danelljan2020probabilistic} & \textcolor{green}{\textbf{0.634}} & \textcolor{blue}{\textbf{0.738}} & \textcolor{blue}{\textbf{0.543}} & 30 \\
  RPT \cite{ma2020rpt} & 0.624 & 0.730 & 0.504 & 20 \\
  Ocean \cite{zhang2020ocean} & 0.611 & 0.721 & - & 58 \\
  DiMP-50 \cite{bhat2019learning} & 0.611 & 0.717 & 0.492 & 43 \\
  D3S \cite{lukezic2020d3s} & 0.597 & 0.676 & 0.462 & 25 \\
  SiamFC++ \cite{xu2020siamfc++} & 0.595 & 0.695 & 0.479 & 90 \\
  SiamCAR \cite{guo2020siamcar} & 0.569 & 0.670 & 0.415 & 52 \\
  ATOM \cite{danelljan2019atom} & 0.556 & 0.634 & 0.402 & 30 \\
  SiamRPN++ \cite{li2019siamrpn++} & 0.517 & 0.616 & 0.325 & 35 \\
  \bottomrule
 \end{tabular}
 \vspace{-1.0em}
\end{table}

\textbf{TrackingNet.}
TrackingNet \cite{muller2018trackingnet} is a large-scale short-term tracking dataset that provides a large amount of videos in the wild for training and testing.
The \textit{testing} set contains 511 videos without publicly released ground truths.
We evaluate our tracker on the \textit{testing} set and obtain results from the dedicated evaluation server.
As shown in \tabref{tab:trackingnet_tab}, our tracker outperforms all previous state-of-the-art real-time approaches by a large margin and strongly sets leading performance.
It is noteworthy that this dataset has a wide variety in terms of classes and scenarios in the wild;
therefore, the significant performance improvement of our approach illustrates its strong generalization ability to real-world tracking videos.
\par
\textbf{GOT-10k.}
GOT-10k \cite{huang2019got} is a recently released large-scale generic object tracking benchmark, containing 10,000 videos totally, in which the \textit{testing} set has 180 videos.
Similar to TrackingNet, the ground truths of the \textit{testing} set are also withheld so that all tracking results must be evaluated in a specific evaluation server.
Different from others, GOT-10k benchmark restricts trackers to use only the \textit{training} set for training.
In this work, we follow this protocol for training our tracker and testing it on the \textit{testing} set.
All settings are unchanged except for the training data.
\tabref{tab:got10k_tab} lists a comparison of our tracker with other state-of-the-art trackers in terms of average overlap (AO) and success rates (SR) at threshold 0.5 and 0.75.
Benefit from the pixel-level similarity computation in the space-time memory reader, our method outperforms the second place tracker PrDiMP-50~\cite{danelljan2020probabilistic} by $3.2\%$ for the $\text{SR}_{0.75}$ metric (The percentage of successfully tracked frames where the overlaps exceed 0.75).
\par
\begin{figure*}[t]
    \centering
    \subfigure[Success Plot]{\includegraphics[width=0.31\textwidth]{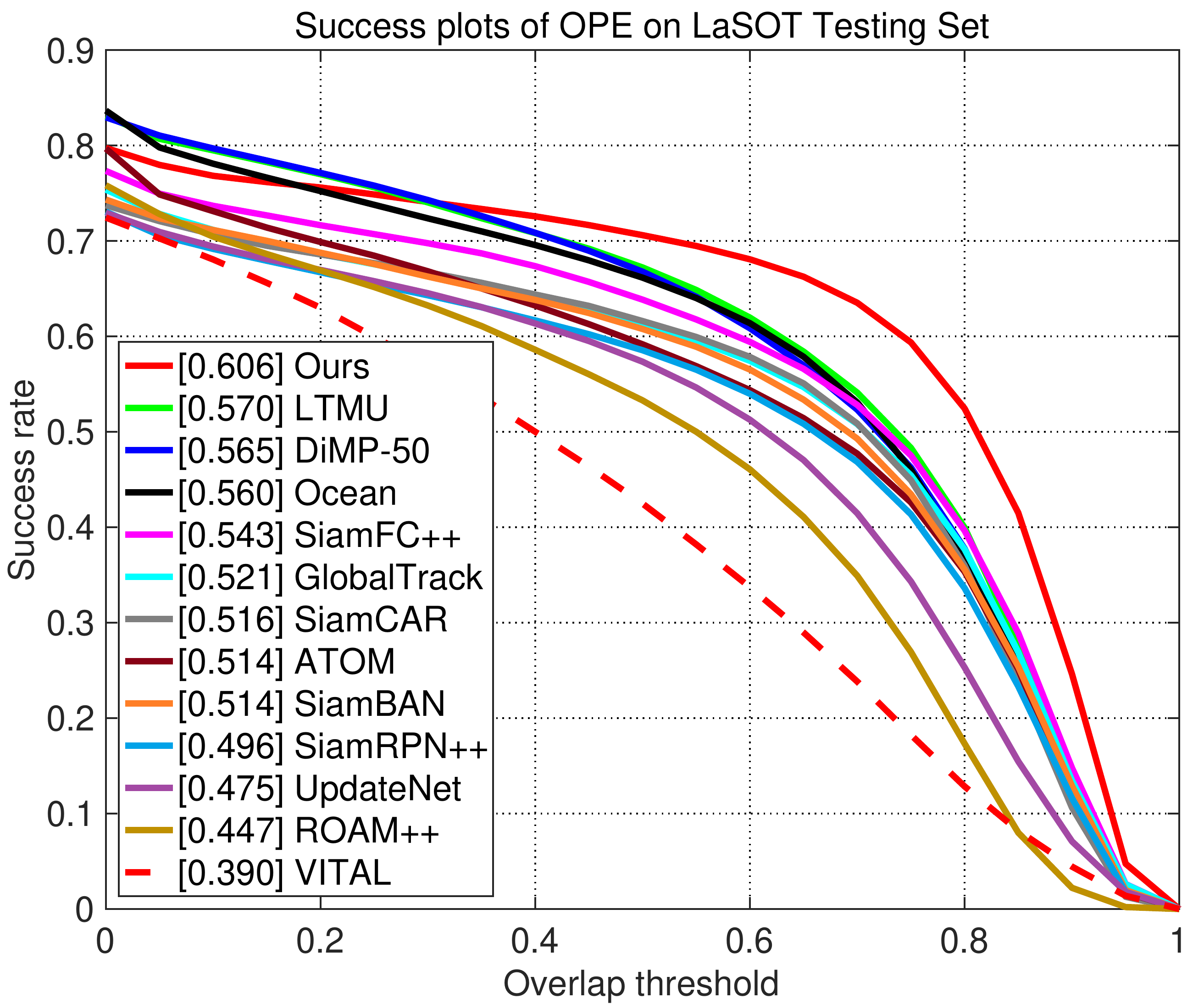}}
    \subfigure[Precision Plot]{\includegraphics[width=0.31\textwidth]{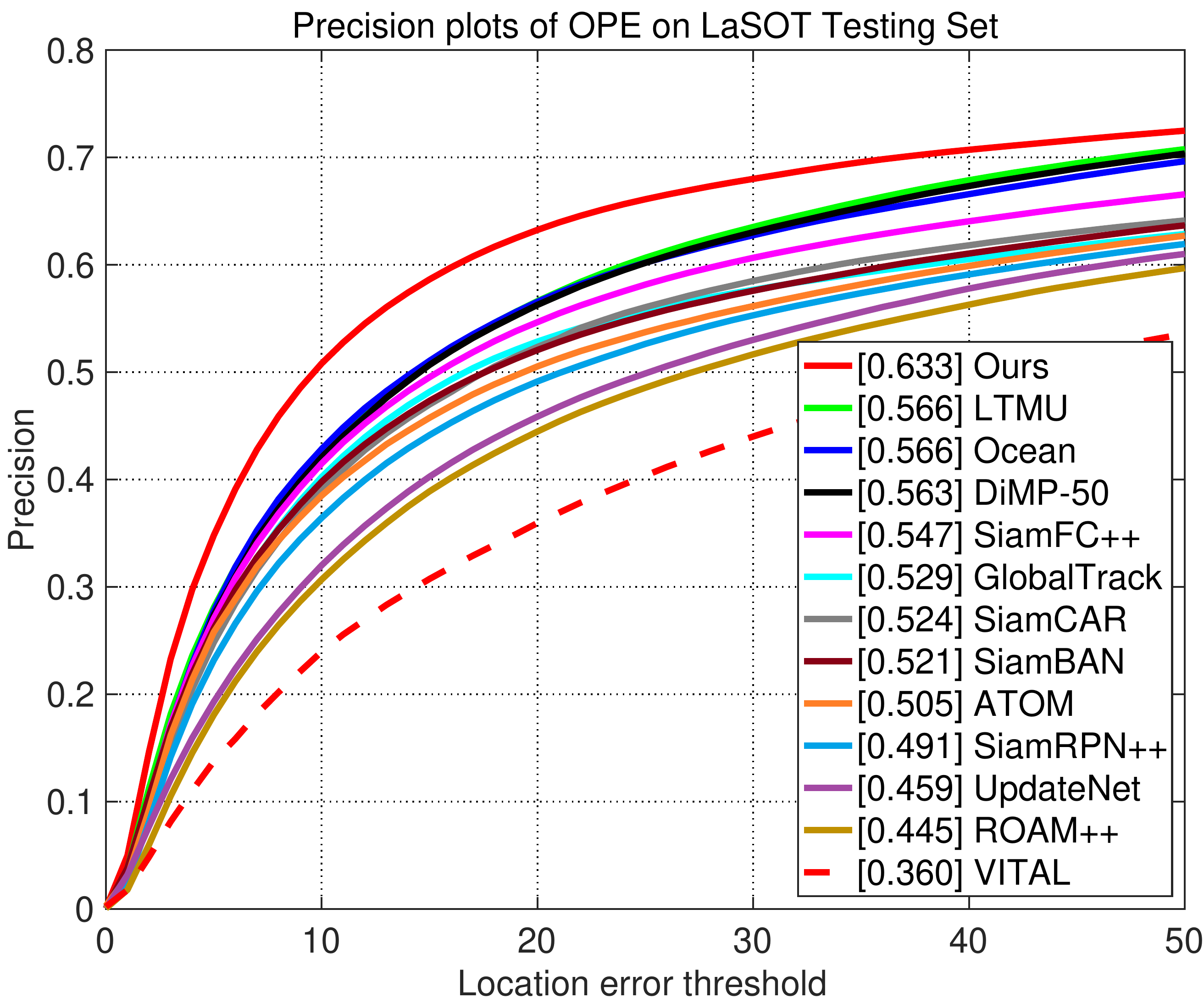}}
    \subfigure[Normalized Precision Plot]{\includegraphics[width=0.31\textwidth]{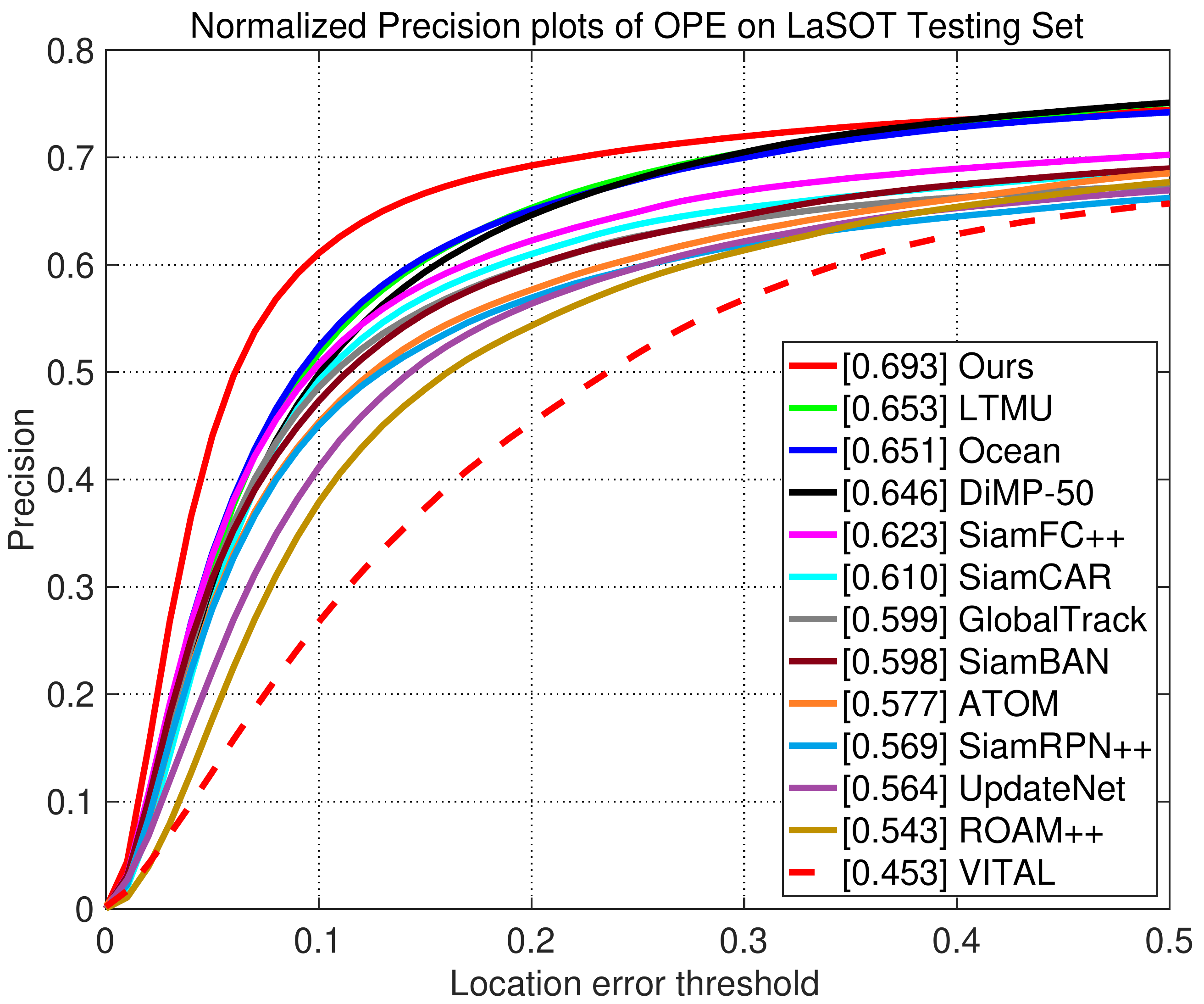}}
    \vspace{-1.0em}
    \caption{Plots show comparisons of our tracker with other competitive trackers on the \textit{testing} set of LaSOT. Trackers are evaluated by the success, precision, and normalized precision metrics.}
    \label{fig:lasot_figure}
\end{figure*}

\textbf{LaSOT.}
LaSOT \cite{fan2019lasot} is also a large-scale single object tracking dataset with high-quality annotations.
Its \textit{testing} set consists of 280 long videos, with an average of 2500 frames per video.
Thus, the robustness of trackers is crucial against complicated scenarios, such as occlusions, out-of-view, \etc.
We report the results on the \textit{testing} set.
As shown in \figref{fig:lasot_figure}, comparing with twelve comparable trackers, our tracker sets top performance in terms of success, precision, and normalized precision.

\begin{table*}[t]
 \centering
 \caption{\label{tab:uav123_tab}A comparison of our tracker with other competitive approaches on UAV123 in terms of success (AUC) metric. The best three results are highlighted in \textcolor{red}{\textbf{red}}, \textcolor{blue}{\textbf{blue}}, and \textcolor{green}{\textbf{green}}, respectively.}
 \vspace{-1.0em}
 \begin{tabular}{c|cccccccc}
  \toprule
  \ & \textbf{Ours} & \textbf{DiMP-50}~\cite{bhat2019learning} & \textbf{ATOM}~\cite{danelljan2019atom} & \textbf{SiamBAN}~\cite{chen2020siamese} & \textbf{SiamCAR}~\cite{guo2020siamcar} & \textbf{SiamRPN++}~\cite{li2019siamrpn++} & \textbf{UPDT}~\cite{bhat2018unveiling}\\
  \midrule
  \textbf{Success} & \textcolor{blue}{\textbf{0.647}}  & \textcolor{red}{\textbf{0.654}}  & \textcolor{green}{\textbf{0.643}}  & 0.631  & 0.614  & 0.613 & 0.545 \\
  \bottomrule
 \end{tabular}
 \vspace{-1.0em}
\end{table*}

\begin{table}[tbp]
 \centering
 \caption{\label{tab:vot2018_tab}A comparison of our tracker with state-of-the-art trackers on VOT2018. The best results are highlighted in \textcolor{red}{\textbf{red}}, \textcolor{blue}{\textbf{blue}}, and \textcolor{green}{\textbf{green}}, respectively. Trackers are ranked from top to bottom according the \textbf{EAO} scores. The arrows after the metrics mean that the bigger($\uparrow$) or the smaller($\downarrow$) is the better.}
 \vspace{-1.0em}
 \begin{tabular}{cccc}
  \toprule
  \textbf{Tracker} & \textbf{EAO$\uparrow$} & \textbf{A$\uparrow$} & \textbf{R$\downarrow$} \\
  \midrule
  \textbf{Ours} & 0.447 & 0.590 & 0.159 \\
  D3S \cite{lukezic2020d3s} & \textcolor{red}{\textbf{0.489}} & \textcolor{red}{\textbf{0.640}} & \textcolor{green}{\textbf{0.150}} \\
  Ocean \cite{zhang2020ocean} & \textcolor{red}{\textbf{0.489}} & 0.592 & \textcolor{red}{\textbf{0.117}} \\
  SiamAttn \cite{yu2020deformable} & \textcolor{blue}{\textbf{0.470}} & \textcolor{green}{\textbf{0.630}} & 0.160 \\
  KYS \cite{bhat2020know} & \textcolor{green}{\textbf{0.462}} & 0.609 & \textcolor{blue}{\textbf{0.143}} \\
  SiamBAN \cite{chen2020siamese} & 0.452 & 0.597 & 0.178 \\
  PGNet \cite{liaopg} & 0.447 & 0.618 & 0.192 \\
  PrDiMP-50 \cite{danelljan2020probabilistic} & 0.442 & 0.618 & 0.165 \\
  DiMP-50 \cite{bhat2019learning} & 0.440 & 0.597 & 0.153 \\
  Siam R-CNN \cite{voigtlaender2020siam} & 0.408 & 0.609 & 0.220 \\
  SiamFC++ \cite{xu2020siamfc++} & 0.426 & 0.587 & 0.183 \\
  SiamRPN++ \cite{li2019siamrpn++} & 0.414 & 0.600 & 0.234 \\
  FCOS-MAML \cite{wang2020tracking} & 0.392 & \textcolor{blue}{\textbf{0.635}} & 0.220 \\
  \bottomrule
 \end{tabular}
 \vspace{-1.0em}
\end{table}

\textbf{UAV123.}
UAV123 \cite{mueller2016benchmark} is designed to evaluate trackers in UAV applications, including 123 low altitude aerial videos, with an average of 915 frames per video.
Due to the characteristics of UAV, this dataset has numerous scenarios with partial and full occlusions, out-of-view, and small objects.
Thus, many objects have quite low resolutions.
As shown in \tabref{tab:uav123_tab}, however, our tracker obtains a success (AUC) score of 0.647, which still significantly outperforms recent competitive siamese trackers SiamBAN~\cite{chen2020siamese}, SiamCAR~\cite{guo2020siamcar}, and SiamRPN++~\cite{li2019siamrpn++}, while running at a real-time speed.
\par
\textbf{VOT2018.}
The 2018 version of the visual object tracking (VOT) challenge \cite{kristan2018sixth} contains 60 videos.
Following the evaluation protocol of the VOT2018 dataset, we report the results of our tracker in terms of expected average overlap (EAO), accuracy (A), and robustness (R) and compare it with state-of-the-art trackers.
As shown in \tabref{tab:vot2018_tab}, the robustness of our tracker is similar to D3S~\cite{lukezic2020d3s} and SiamAttn~\cite{yu2020deformable}.
However, the accuracy is worse than them since the ground truths in the VOT evaluation system are rotated bounding boxes, the estimated bounding boxes in our tracker are axis-aligned instead.

\section{Conclusions}
This work proposes a novel tracking framework based on space-time memory networks.
The framework abandons the traditional template-based tracking mechanism, using multiple memory frames and foreground-background label maps to locate the target in the query frame.
In the space-time memory networks, the target information stored in multiple memory frames is adaptively retrieved by the query frame, so that the tracker has a strong adaptive ability to the target variations.
Extensive experiments demonstrates that, without bells and whistles, the proposed tracker achieves better performance than current state-of-the-art real-time methods, while running at 37 FPS.
The experiments also shows its generalizability, extendibility, and applicability.

{\small
\bibliographystyle{ieee_fullname}
\bibliography{egbib}
}
\end{document}